\documentclass[10pt,twocolumn,letterpaper]{article}

\usepackage{cvpr}              % To produce the CAMERA-READY version
\usepackage[dvipsnames]{xcolor}

\definecolor{cvprblue}{rgb}{0.21,0.49,0.74}
\usepackage[pagebackref,breaklinks,colorlinks,citecolor=cvprblue]{hyperref}
\usepackage{adjustbox}

\def\paperID{12} % *** Enter the Paper ID here
\def\confName{CVPR}
\def\confYear{2024}

\title{Model editing for distribution shifts in uranium oxide morphological analysis}
\author{Davis Brown\thanks{Equal contribution.}, Cody Nizinski\footnotemark[1], Madelyn Shapiro, 
\\ Corey Fallon, Tianzhixi Yin, Henry Kvinge, and Jonathan H. Tu \\
Pacific Northwest National Laboratory\\
{\tt\small {\ davis.brown, cody.nizinski, jonathan.tu }\ @pnnl.gov}
}
\begin{document}
\maketitle

\begin{abstract}
Deep learning still struggles with certain kinds of scientific data. Notably, pretraining data may not provide coverage of relevant distribution shifts (e.g., shifts induced via the use of different measurement instruments). We consider deep learning models trained to classify the synthesis conditions of uranium ore concentrates (UOCs) and show that \textbf{model editing} is particularly effective for improving generalization to distribution shifts common in this domain. In particular, model editing outperforms finetuning on two curated datasets comprising of micrographs taken of U$_{3}$O$_{8}$ aged in humidity chambers and micrographs acquired with different scanning electron microscopes, respectively.
\end{abstract}

\section{Introduction}

The morphological features of nuclear materials can provide information about the processing conditions used to produce the materials and the history of the materials after their production \cite{mcdonald2023review}. For uranium ore concentrates (UOCs) the precipitation chemistry and calcination conditions have been shown to impart certain morphological features that allow particles observed by scanning electron microscopy (SEM) to be correlated back to the synthesis conditions \cite{schwerdt2019uranium}. Aging studies have looked at how UOC properties and particle morphologies change over time during exposure to different storage conditions, particularly humidity and temperature \cite{sweet2013investigation, olsen2018quantification, tamasi2017morphology, tamasi2015oxidation, wilkerson2020hydration, pastoor2020understanding, hanson2021effect, hanson2021impact}.

A number of analytical methods have been developed for morphological analysis of nuclear materials, including qualitative descriptions \cite{tamasi2016lexicon}, quantitative measurement of particles by image segmentation \cite{olsen2017quantifying} or other particle sizing methods, texture analysis \cite{fongaro2016application, fongaro2021development}, and machine learning. Deep learning models for morphological analysis have been trained in supervised \cite{ly2020determining}, self-supervised \cite{johnson2024improving}, and unsupervised settings \cite{girard2021uranium}. The robustness of these models to out-of-distribution (OOD) data has been investigated \cite{nizinski2022characterization}, along with methods for uncertainty quantification (UQ) or calibration of prediction confidence \cite{hagen2022dbcal} and source-free domain adaptation (SFDA) for covariate shifts \cite{ly2improving}.

Computer vision models for predicting the provenance of nuclear materials must be robust to statistic-level variations (i.e., the appearance of images taken with various microscopes and in varying settings), and to some extent, feature-level variations (i.e., changes to the actual morphological characteristics). Whereas SFDA has only been demonstrated for the former, model editing methods \cite{bau2020rewriting, mitchell2021fast} could potentially be used for both types of distribution shifts and can be applied to existing validated models incrementally as new kinds of distribution shifts are discovered with lower computational needs than other methods.

\subsection*{Contributions}
Our contributions can be summarized as follows:
\begin{itemize}
    \item Model editing methods generalize well on feature-level distribution shifts caused by aging uranium oxides under diel cycling humidity and temperature conditions.
    \item We compare editing methods and find that low-rank editing generally outperforms surgical finetuning. 
    \item Both editing and surgical finetuning outperform full-model finetuning.
\end{itemize}

\section{Description of data}

The training and validation datasets consist of scanning electron microscope (SEM) images collected of UOCs representing five precipitation pathways --- ammonium diuranate (ADU),  ammonium uranyl carbonate (AUC), magnesium diurante (MDU), sodium diuranate (SDU), and uranyl peroxide (UO$_{4}\cdot$H$_{2}$O) --- that have been converted to three uranium oxides --- uranium trioxide (UO$_{3}$), triuranium octoxide (U$_{3}$O$_{8}$), and uranium dioxide (UO$_{2}$). The micrographs were collected using an FEI Nova NanoSEM 630 scanning electron microscope with the through lens detector (TLD) operating in secondary electron (SE) mode. More complete descriptions of the material synthesis and data collection can be found elsewhere \cite{schwerdt2019uranium, ly2020determining, heffernan2019identifying, abbott2022thermodynamic}. Previous work has shown that classification models trained on earlier versions of this dataset fail to generalize to new data that differs by process history and/or imaging parameters \cite{nizinski2022characterization}.

\begin{figure*}[!ht]
\centering
\centering
\includegraphics[width=\textwidth]{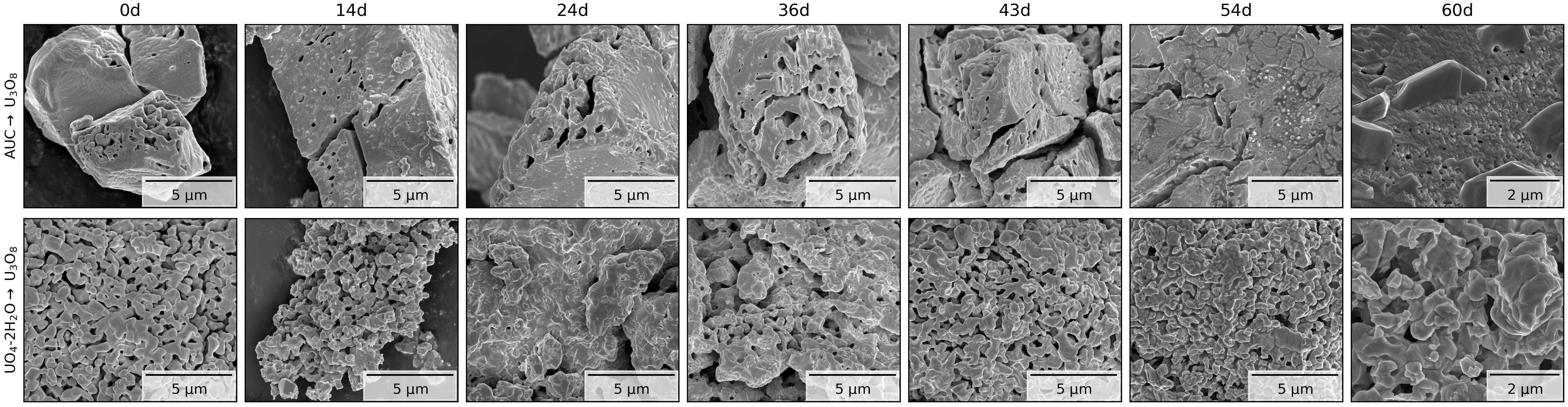}%
\caption{Images from aging datasets. Note the (visually apparent) differences in the 60 day aged samples from the others.}
 \label{fig:agingexamples}
\end{figure*}

\begin{figure*}[!ht]
\centering
\includegraphics[trim=0 0 0 0.51cm, clip, width=\textwidth]{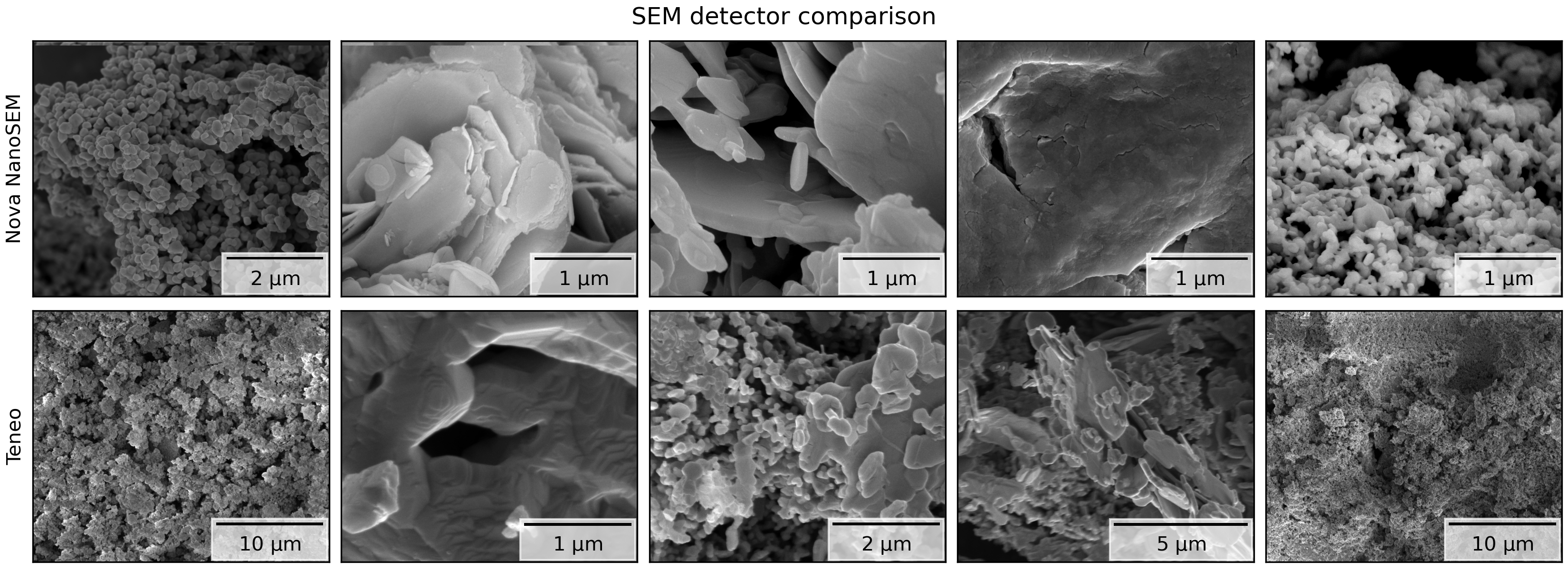}%
\caption{SEM detector comparison images from Nova NanoSEM and Teneo models.}
\label{fig:detectorexamples}
\end{figure*}

We consider two different concept distribution shifts, an \textbf{aging} shift and a \textbf{detector} shift: 
\begin{itemize}
    \item The dataset for the \textbf{aging} editing task uses micrographs of materials that were initially U$_{3}$O$_{8}$ produced by calcining either AUC or UO$_{4}\cdot$2H$_{2}$O. The starting U$_{3}$O$_{8}$ was then aged under conditions that cycled between high temperature/humidity and low temperature/humidity following a 24-hour daily cycle; samples were pulled at time steps of $\{0, 14, 36, 43, 54, 60\}$ days to collect morphology data by SEM and measure crystallographic changes by X-ray diffraction (XRD) \cite{hanson2021effect}. We provide examples in \cref{fig:agingexamples}.
    \item The \textbf{detector} editing task uses micrographs of the same materials as the train and validation sets but collected on a different SEM, a FEI Teneo SEM with the ``T2'' SE detector \cite{ly2improving}. Example images in \cref{fig:detectorexamples}.
\end{itemize}
\cref{tab:dataset_counts} provides the number of images in our train and validation splits for the base dataset (for training our model) and the concept datasets (for the editing task).

\begin{figure}[!htb]
  \centering
  \begin{subfigure}[b]{0.4\textwidth}
    \centering
      \includegraphics[width=\textwidth]{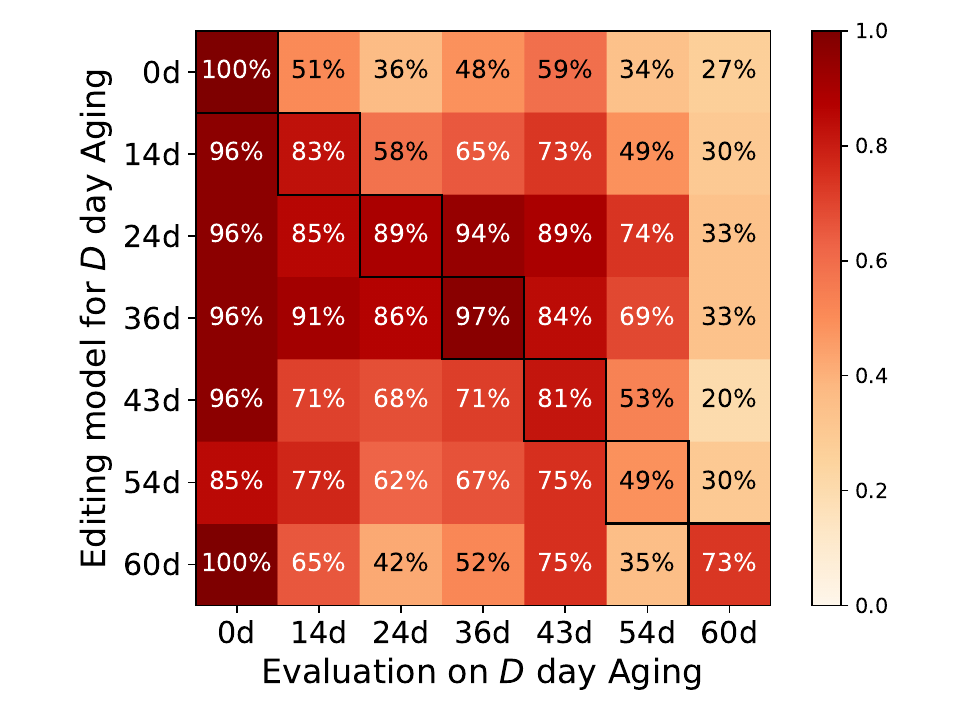}%
    \caption{Surgical finetuning accuracy}
    \label{fig:agingplot1}
  \end{subfigure}
  \hfill
  \begin{subfigure}[b]{0.4\textwidth}
    \centering
      \includegraphics[width=\textwidth]{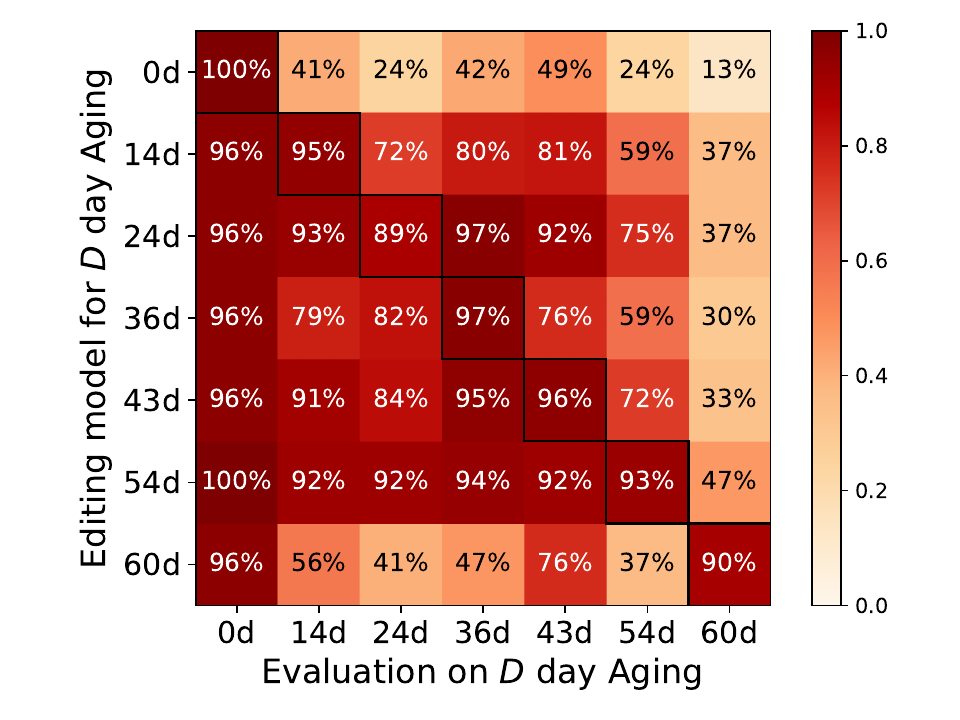}%
    \caption{Low-rank editing accuracy}
    \label{fig:agingplot2}
  \end{subfigure}
  \hfill
  \begin{subfigure}[b]{0.4\textwidth}
    \centering
    \includegraphics[width=\textwidth]{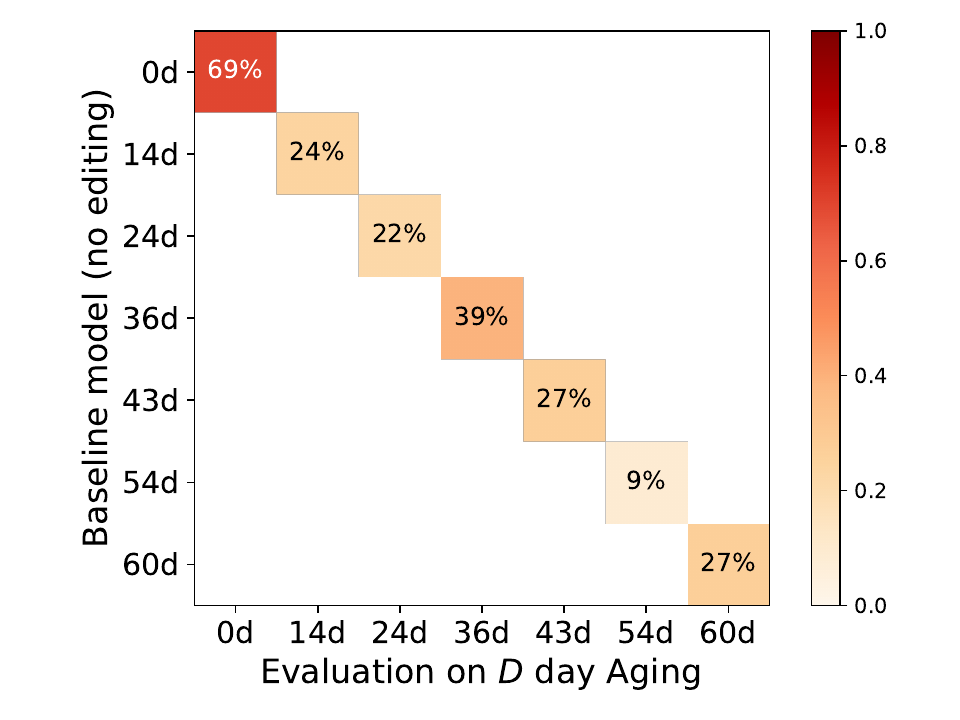}
    \caption{Baseline model (unedited) accuracy}
    \label{fig:agingplot3}
  \end{subfigure}
  \caption{Comparison of how well models finetuned on an aging dataset $D$ generalize to other ages. Both surgical finetuning (\textbf{(a)}) and low-rank editing (\textbf{(b)}) generalize to earlier ages. Also, all edited models outperform the baseline accuracy in \textbf{(c)}). Note that because we drop runs that incur an accuracy drop on the baseline validation set of $>1.5\%$, there are no succesful full-model finetuning runs. See \cref{fig:agingplots2} for a more permissive threshold.}
  \label{fig:agingplots}
\end{figure}

\section{Methods for model updating}
Model editing makes tweaks to model weights to incorporate new facts \citep{mitchell2021fast, meng2022locating, hase-etal-2023-methods}, semantics \citep{bau2020rewriting}, and behaviors \citep{ilharco2022editing}. Such methods have been used to patch model errors in the form of spurious correlations \citep{NEURIPS2021_c46489a2} and confusion \citep{Panigrahi_2023_CVPR} with only a single input example. We focus on the setting of \textit{natural distribution shifts}, where we want to adapt models to inherent variations in the data --- largely in the form of input/statistical-level variations (the \textbf{detector} editing task) and feature-level variations (the \textbf{aging} editing task).

Let $f$ be a neural network defined as a composition of layers $f=f_L \circ f_{L-1} \circ \cdots \circ f_2 \circ f_1$, where $f_l(x)=\sigma\left(W_l x\right)$ and $W_l$ is an $n_l \times n_{l-1}$ matrix for each $l=1, \ldots, L$ with an activation function $\sigma$. 
For any layer, let $f_{\leq l}$ denote the composition of the first $l$ layers: $f_l \circ f_{l-1} \circ \cdots \circ f_2 \circ f_1$.
In contrast with other work, e.g. \citep{bau2020rewriting, meng2022locating, Panigrahi_2023_CVPR}, we do not use exemplar pairs $(x, x^{\prime}, y)$ of transformed images $x^{\prime}$ with a fixed label $y$ to update a model so that $f_{\leq l}(x) \approx f_{\leq l}\left(x^{\prime}\right)$. 
Instead, we update our model on new pairs of datapoints $x^{\prime}$ and labels $y^{\prime}$ from either the aging shift or the detector shift and use stochastic gradient descent (SGD) so that $f(x') \approx y'$. 
This is a fairly significant difference from some of the editing methods discussed above, however we keep the editing terminology because our methods modify single-layers with constrained updates and/or a limited number of examples.

We next describe the two model update methods we use in this work. 
We compare these update methods to a baseline of \textit{full finetuning}, where all model weights $W_1, \ldots, W_L$, are updated with stochastic gradient descent (SGD) to minimize the mean squared error (MSE) of $f(x^{\prime})$ and $y^{\prime}$.

\noindent \textbf{Low-rank model editing}: the edit is an update $W_l \leftarrow W_l+U V^T$ where $U$ and $V$ are low-rank matrices learned via SGD to minimize the MSE of $f(x^{\prime})$ and $y^{\prime}$; in our experiments we fix the rank to $r=2$.
The model weights $W_1, \ldots, W_L$, are kept frozen. 
We use the implementation described in \citep{bau2020rewriting} where $U$ and $V$ are $1 \times 1$ convolutions when applied to the weights of convolutional layers.

\noindent \textbf{Surgical finetuning}: we largely follow \citep{lee2022surgical}, updating only a subset of layers with SGD to minimize the MSE of $f(x^{\prime})$ and $y^{\prime}$.
 In order to make a more precise comparison with low-rank model editing, we deviate from \citep{lee2022surgical} and optimize only a single layer, $W_l$, rather than blocks of layers. This is also called \textit{local finetuning} in \citep{bau2020rewriting}.

\section{Experiments}

For both datasets, our goal is to adapt the model for the new domain while maintaining the model's performance on the original dataset. 
On the aging dataset, this means updating a model to be accurate on the set of images and labels $(x^{\prime}, y^{\prime})$ that have undergone the $D$ 24-hour cycles of aging, while maintaining performance on the original unaged datapoints $(x, y)$.
Similarly, for the detector dataset the model is updated to be accurate on datapoints taken from both the T2 SE detector and the original detector.

We use a ConvNeXt-Small model \citep{liu2022convnet} trained with SGD on the original SEM image dataset with a validation accuracy of $97.9\%$. For each of the model updating methods, we perform a coarse grid search over learning rates for each convolutional layer, and then a finer-grained grid search over the two best performing layers for each method. To choose the best performing hyperparameters for an update, we employ the  hyperparameter selection strategy from \citep{bau2020rewriting}, where we select $50\%$ of the dataset examples (for a given age $D$ for the aging dataset or for the images taken from the T2 SE detector) to learn the update and to select the best hyperparameters and the remaining $50\%$ to test performance. A hyperparameter run is dropped if it causes the model performance on the original SEM validation set to drop below a given threshold (e.g., a threshold of $1.5\%$ drop for \cref{fig:agingplots} and a substantially more permissive $7\%$ threshold for \cref{fig:agingplots2}). 

\subsection{Aging Experiments}
\begin{figure*}[!ht]
  \centering
  \begin{subfigure}[b]{0.33\textwidth}
    \centering
      \includegraphics[width=\textwidth]{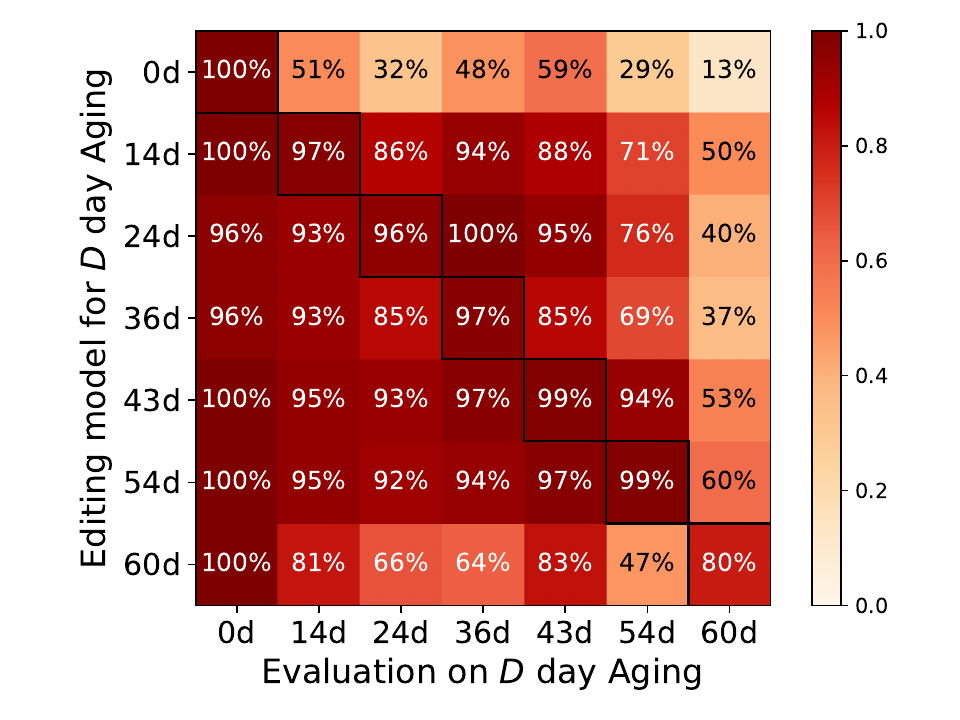}%
    \caption{Surgical finetuning accuracy}
    \label{fig:aging2plot1}
  \end{subfigure}
  \hfill
  \begin{subfigure}[b]{0.33\textwidth}
    \centering
      \includegraphics[width=\textwidth]{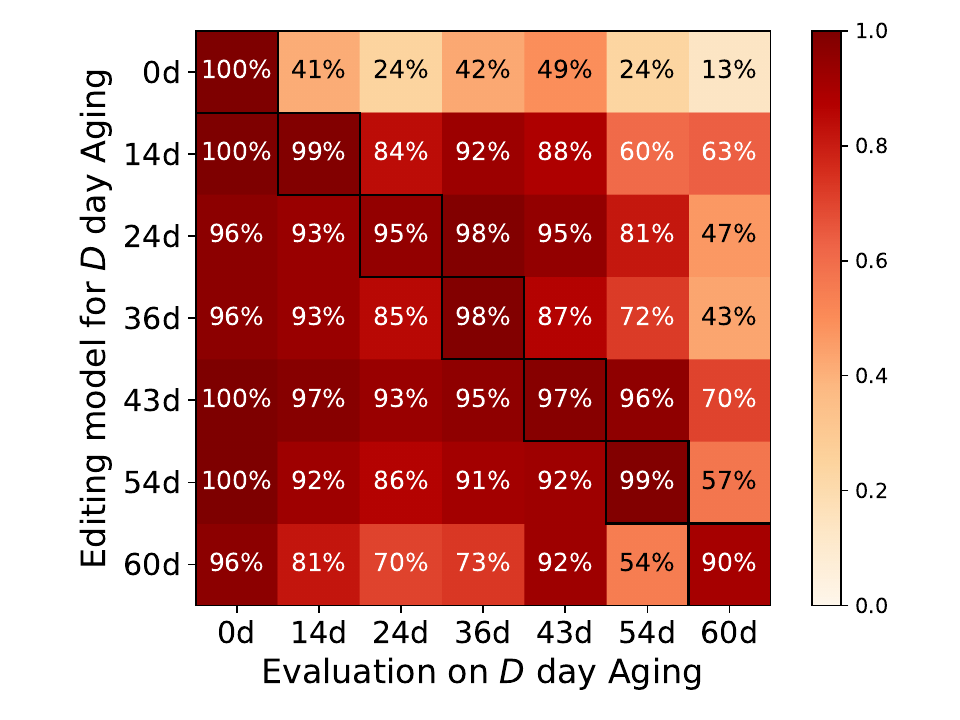}%
    \caption{Low-rank editing accuracy}
    \label{fig:aging2plot2}
  \end{subfigure}
  \hfill
  \begin{subfigure}[b]{0.33\textwidth}
    \centering
    \includegraphics[width=\textwidth]{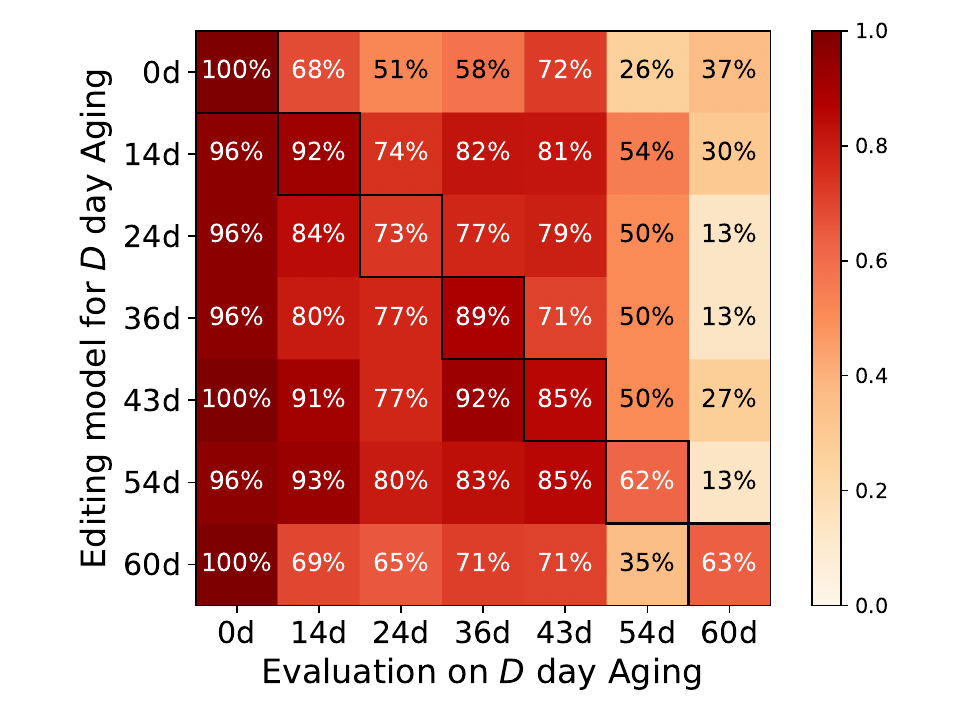}
    \caption{Full finetuning accuracy}
    \label{fig:aging2plot3}
  \end{subfigure}
  \caption{Comparison for how well models updated for an aging dataset $D$ generalize to other ages, allowing up to a $7\%$ drop in accuracy on the original validation set. Even for this more aggressive editing setting (as opposed to \cref{fig:agingplots}, which used a threshold of $1.5\%$), surgical finetuning and low-rank editing tend to outperform full finetuning. The aging accuracies for the unedited model are given in  \cref{fig:agingplot3}.}
 . \label{fig:agingplots2}
\end{figure*}
The aging editing task comprises seven different material aging time steps.
We use this dataset structure to evaluate how well models edited with images of age $D$ generalize to other durations of particle aging. 
In other words, we ask how well a model update captures the relevant \textit{morphology} of aging as opposed to the \textit{particulars} of the collected dataset/aging timing.
For example, we examine how well a material aged for $D=43$ days performs on unaged control samples ($0$ days), earlier ages (aging for $14, 24,$ and $36$ days, respectively) and later ages ($54$ and $60$ days). 

\cref{fig:agingplots} summarizes the results of the aging experiments.
The diagonal elements capture the performance of a model on the targeted aging duration whereas off-diagonal elements capture the generalization of those models to other aging durations.
The performance of the original model (without any updates) is plotted in \cref{fig:agingplot3}.
Because all full-model finetuning runs incurred substantial accuracy drops on the base (non-aged) validation set, we also plot performance for a more permissive threshold in \cref{fig:agingplots2}.
Note that all of the model update methods --- surgical finetuning,  low-rank editing, and even the baseline full-model finetuning for the permissive threshold --- exhibit partial generalization across different ages.
The generalization to shorter aging durations is stronger than for longer ones.

Comparing rows of \cref{fig:agingplot1} to those of \cref{fig:agingplot2}, we see that low-rank editing outperforms surgical finetuning on the targeted aging duration (e.g., editing a model on 43-day aging examples and testing on held-out 43-day aging examples) as well as on other aging datasets (e.g., editing a model on 43-day aging examples and testing on hold-out data for $D \neq 43$).
For example, in the aging tasks for $43$ days and $54$ days, low-rank editing has better editing performance for the target editing task by $15\%$ for $43$ days and $44\%$ for $54$ days of aging.
For the off-target editing generalization to other aging sets, low-rank editing outperforms surgical fine-tuning as much as $27\%$.
Both update methods increase model aging accuracy over the baseline model in \cref{fig:agingplot3} and over full finetuning in \cref{fig:agingplots2}, and are targeted, i.e., do not pay substantial performance costs on the original SEM dataset.
We expect that the better targeting of editing is due to regularization, where we update only a single-layer for surgical finetuning and single-layer with a low-rank restriction for low-rank editing.

Finally, we note that all update methods have worse generalization on the 60-day aging data. While the updates on the 60-day aging duration achieve reasonable performance on the held-out 60-day set, they do not generalize well to $D\neq60$.
Likewise, updating a model for $D\neq60$ leads to relatively minimal increases in accuracy for $D=60$. 
We hypothesize that this poor performance is due to slightly different collection conditions for the 60-day micrographs, where a different scientist collected the SEM samples. 
This suggests that the success of model updates is sensitive to the image collection parameters; rather than editing for the changes in surface morphology seen in $D=60$ materials, edits were made on the style (e.g., brightness, contrast) of the image. 
We expect that editing performance will increase when using editing prototypes that better reflect the diversity in SEM images, however we leave this to future work.

%In particular, rather than editing for `changes in morphology due to material aging' here we edit for `changes in morphology due to material aging under our collection parameters.'

\subsection{SEM Detector Experiments}
\begin{figure}[!ht]
  \centering
  \begin{subfigure}[b]{0.45\textwidth}
    \centering
      \includegraphics[width=\textwidth]{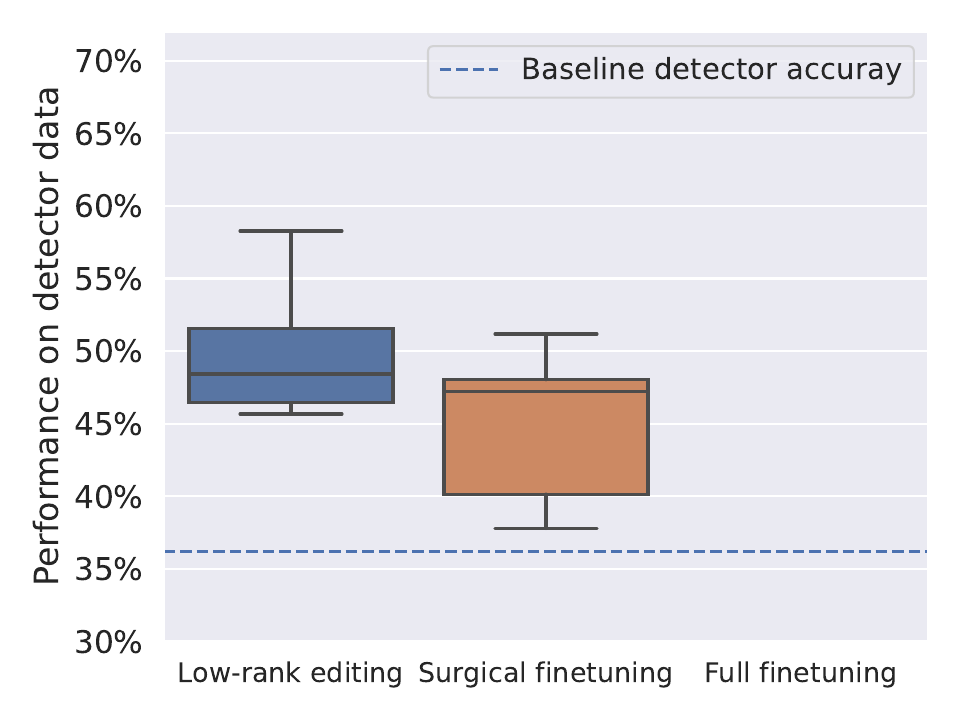}%
    \caption{Detector accuracy, removing models with an original validation accuracy drop greater than $1.5\%$}
    \label{fig:boxplot1}
  \end{subfigure}
  \begin{subfigure}[b]{0.45\textwidth}
    \centering
      \includegraphics[width=\textwidth]{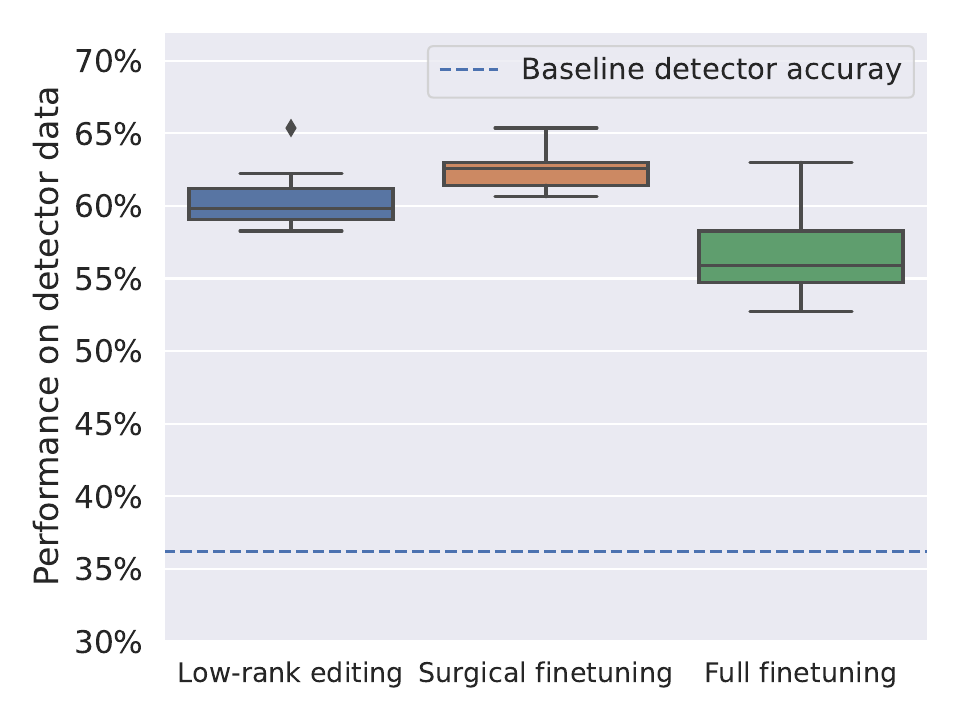}%
    \caption{Detector accuracy, removing models with an original validation accuracy drop greater than $7\%$.}
    \label{fig:boxplot2}
  \end{subfigure}
  \hfill
  \caption{Model editing performance on the T2 SE detector dataset.}
  \label{fig:detectorbox}
\end{figure}

The detector editing task captures input-level and statistical changes in the distribution of SEM images due to using a different scanning electron microscope.
In general, this task is more challenging for the editing methods under consideration.
While editing does provide performance improvement over the baseline unedited model, the accuracy increases are modest.
Box plots of model accuracy across five random seeds for each hyperparameter are plotted in \cref{fig:detectorbox}.
Like in the aging task, we learn model updates for the detector task with $50\%$ of the SEM images, test on the remaining $50\%$, and drop runs that incur an accuracy drop on the original validation above a certain threshold (dropping runs with performance drops greater than $1.5\%$ for \cref{fig:boxplot1} and $7\%$ for \cref{fig:boxplot2}).

For the more restrictive validation dropping threshold of $1.5\%$ in \cref{fig:boxplot1}, low-rank editing tends to outperform surgical finetuning. Like in the aging editing task runs in \cref{fig:agingplots}, there were no training runs from full finetuning that met this threshold. With the more permissive threshold of $7\%$ in \cref{fig:boxplot2}, surgical finetuning and low-rank editing again generally outperform full finetuning.

\section{Conclusions and Future Work}
We find that we can update models trained to predict the processing conditions of uranium ore concentrates to better handle a common domain-specific distribution shifts. 
Model editing methods perform particularly well on the set of aging distribution shifts, representing feature-level morphological changes in the SEM images.
This has the potential to be particularly useful for designing future aging studies and incorporating their data, where model edits for a duration $D$ can be reliably expected to generalize to shorter durations (and to a lesser but still notable extent, longer ones).

The detector distribution shift, reflecting lower-level variation in the statistics of input samples, proved more difficult. We are particularly optimistic about two directions for future work. 
First, we think that more targeted mixtures of image examples used for editing may better capture relevant directions of variation. For example, using exemplars from multiple detectors may allow for better generalization on a held-out detectors. 
Second, strong results in this domain have been achieved by using generative models to aide domain adaptation \citep{ly2improving}. 
Related to domain adaptation, recent work has leveraged generative models to automatically discover failure modes \citep{wiles2022discovering} and debug them \citep{chegini2023identifying}. 
Such approaches may find a favorable balance between the domain coverage of generative methods with the fine-grained targeting of model editing.

\section{Acknowledgements}
This research was supported by the Mathematics for Artificial Reasoning in Science (MARS) initiative at Pacific Northwest National Laboratory.
It was conducted under the Laboratory Directed Research and Development (LDRD) Program at at Pacific Northwest National Laboratory (PNNL), a multiprogram
National Laboratory operated by Battelle Memorial Institute for the U.S. Department of Energy under Contract
DE-AC05-76RL01830.

{
    \small
    \bibliographystyle{ieeenat_fullname}
    \bibliography{main}
}

\newpage

\section*{Appendix}

\subsection{Dataset Splits}

\begin{table}[ht]
\centering
\begin{tabular}{|l|c|}
\hline
\textbf{Dataset} & \textbf{Number of Images} (Train / Val) \\
\hline
Base Dataset & 3862 / 965 \\
Detector Dataset & 126 / 127 \\

\hline
\multicolumn{2}{|c|}{Aging Datasets} \\
\hline
Aging 0d Dataset & 25 / 26 \\
Aging 14d Dataset & 74 / 75 \\
Aging 24d Dataset & 73 / 74 \\
Aging 36d Dataset & 66 / 66 \\
Aging 43d Dataset & 74 / 75 \\
Aging 54d Dataset & 68 / 68 \\
Aging 60d Dataset & 30 / 30 \\

\hline
\end{tabular}
\caption{Number of Images in Base and Concept Datasets. We use an 80/20 split for pretraining the original base dataset and a 50/50 train/validation split for our editing methods.}
\label{tab:dataset_counts}
\end{table}

\subsection{Concept dataset examples}

\end{document}